\documentclass{article}
\usepackage{spconf,amsmath,graphicx}
\usepackage{booktabs}
\usepackage{amssymb, amsmath, amsfonts}
\usepackage[linesnumbered,ruled]{algorithm2e}
\usepackage[colorlinks,urlcolor=red,citecolor=blue]{hyperref}

\usepackage[T1]{fontenc}
\usepackage{beramono}
\usepackage{enumitem}

\title{Disentangling Controllable Object through Video Prediction Improves Visual Reinforcement Learning}
\name{Yuanyi Zhong \qquad Alexander Schwing \qquad Jian Peng}
\address{University of Illinois at Urbana-Champaign, IL, USA}
			    
\begin{document}
%
\maketitle
%

\begin{abstract}
In many vision-based reinforcement learning (RL) problems, the agent controls a movable object in its visual field, e.g., the player's avatar in video games and the robotic arm in visual grasping and manipulation. Leveraging action-conditioned video prediction, we propose an end-to-end learning framework to disentangle the controllable object from the observation signal. 
The disentangled representation is shown to be useful for RL as additional observation channels to the agent. Experiments on a set of Atari games with the popular Double DQN algorithm demonstrate improved sample efficiency and game performance (from $222.8\%$ to $261.4\%$ measured in normalized game scores, with prediction bonus reward). 
\end{abstract}

\begin{keywords}
reinforcement learning, video prediction, representation learning, sample efficiency
\end{keywords}
%

\section{Introduction}
\label{sec:intro}

Recent advances in deep reinforcement learning (RL) have significantly improved the state-of-the-art of several challenging visual decision making and control tasks, such as robot grasping \cite{quillen2018deep, ebert2018visual}, and video game playing in Atari games \cite{mnih2015human}, StarCraft II \cite{alphastarblog} and Dota 2 \cite{openaifive}. 
Modern deep RL solutions typically adopt a deep neural network which directly maps raw sensory input signal to either action probabilities (as in policy optimization) or values (as in valued-based RL).

\begin{figure}[t]
	\centering
	\includegraphics[width=\linewidth]{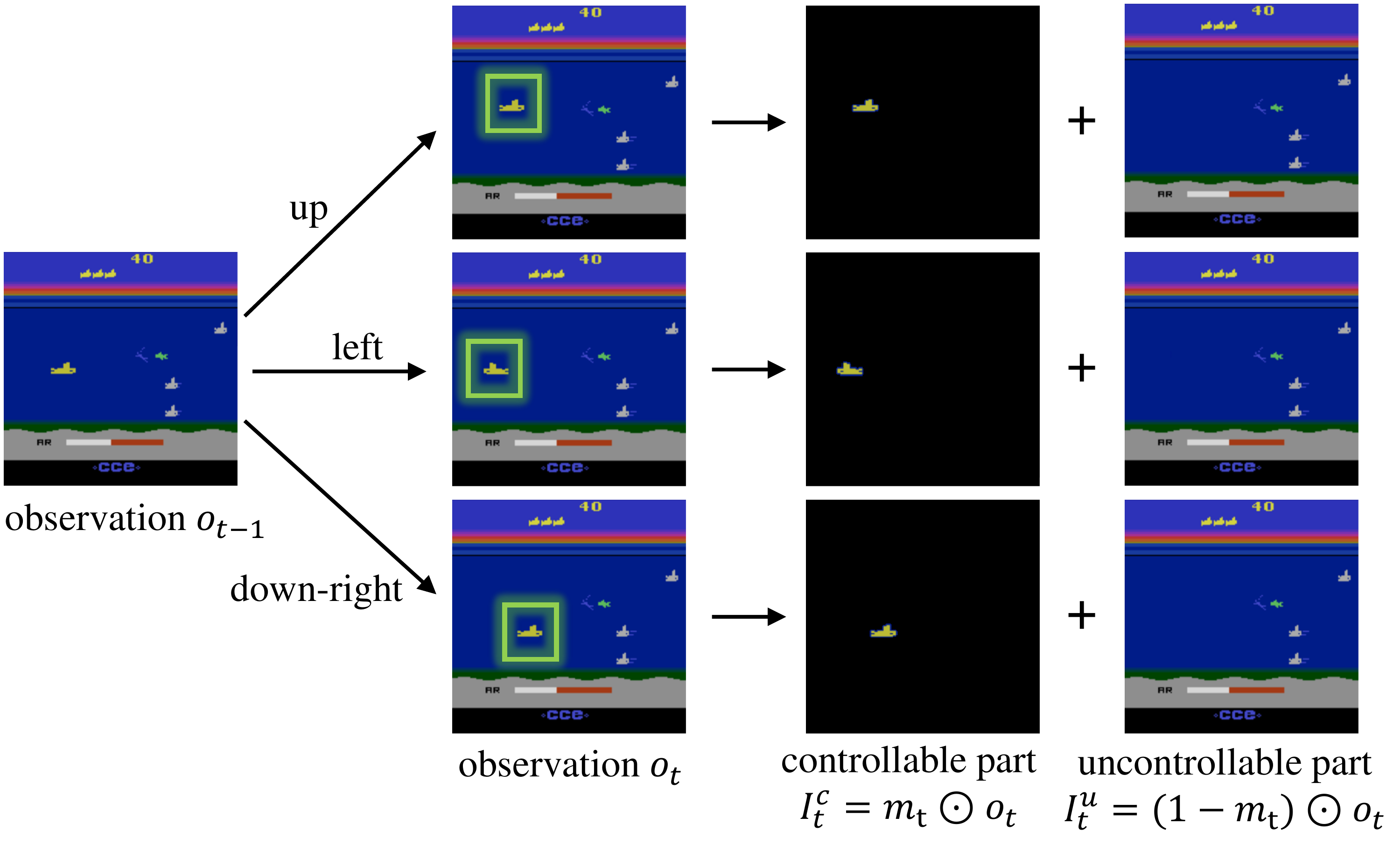}
	\caption{An illustration of controllable object disentanglement in the Atari game Seaquest. The location of the submarine on the future frame depends on the agent's action, whereas the environmental objects are not affected by actions.}
	\label{fig_intro}
	\vspace{-5mm}
\end{figure}

It has been postulated that behind the success of deep learning is its ability to learn good representations \cite{lecun2015deep}, and furthermore, a good representation for signal processing should disentangle the factors of variations \cite{thomas2017independently}. We argue in this paper that the agent's controllable object is an important factor to be isolated from the observation space in vision-based (visual) RL.
Actually, the existence of such controllable objects is prevalent in both real and artificial worlds. In many real-world vision-based control problems, such as autonomous driving and robotic arm manipulation \cite{quillen2018deep}, 
there is typically an object in the visual field that the agent can control. 
While in computer games, the player often controls an avatar on screen that follows the player's commands, as illustrated in Fig.~\ref{fig_intro}.

To achieve the goal of disentangling the controllable objects, we propose a self-supervised learning approach based on action-conditional future frame prediction. By carefully designing the network architecture, the training objective and the regularization terms, the model can be effectively trained without any ground truth annotations, at the same time as the agent's policy learning.

The disentanglement of controllable object is useful for reinforcement learning. We applied our approach on a suite of Atari 2600 games with the Double DQN \cite{DBLP:conf/aaai/HasseltGS16} algorithm, where the agent's input was augmented with the controllable object images. We first verified that our model was able to successfully discover the controllable object regions in those games. Second, the augmented agents achieved both better sample efficiency and higher final test scores, compared to the vanilla DDQN agents which take in only original game frames. Further improvement was obtained by utilizing the prediction error as a bonus reward.

Our contributions are summarized as follows:
\begin{enumerate}[noitemsep, topsep=0pt, leftmargin=1.5em]
\item 
Formulating a video prediction model specifically for visual RL to disentangle the controllable objects in a self-supervised manner; 
\item
Introducing novel training techniques to learn the prediction model;
\item
Incorporating the prediction model and the disentangled representation with DDQN and demonstrating significant improvement on several Atari games.
\end{enumerate}


\section{Frame Prediction Model}



The problem of disentangling controllable image region is formalized as judging whether a pixel $(x,y)$ belongs to an object controlled by the agent on a target frame $o_t$ at time $t$. 
Denoting $m_t(x,y)$ as the probability that pixel $(x,y)$ belongs to the controllable region, then $m_t$ is essentially an attention mask on image $o_t$. 
The frame $o_t$ is then decomposed into two non-intersecting parts as in Eq.~\ref{eq_decomp}: the element-wise product $m_t \odot o_t$ gives the controllable object image region, and $(1-m_t) \odot o_t$ gives the rest uncontrollable regions.
\begin{equation}
\label{eq_decomp}
o_t(x,y) = m_t(x,y) o_t(x,y) + (1-m_t(x,y)) o_t(x,y) .
\end{equation}



The key idea behind our approach is to utilize action-conditioned future frame prediction, and use action information as the bottleneck supervision signal to learn the disentanglement. 
Specifically, we divide the prediction model into three branches: (1) one to predict the action-related image $m_t \odot o_t$ given the \emph{action} $a_t$ and previous frames $o_{t-4}..o_{t-1}$; (2) one to predict the rest of the image $(1-m_t) \odot o_t$ given \emph{only} the previous frames; (3) a mask image prediction $m_t = m(o_t)$ given frame $o_t$ to combine branches (1) and (2). 


\begin{figure}[bt]
	\centering
	\includegraphics[width=0.7\linewidth]{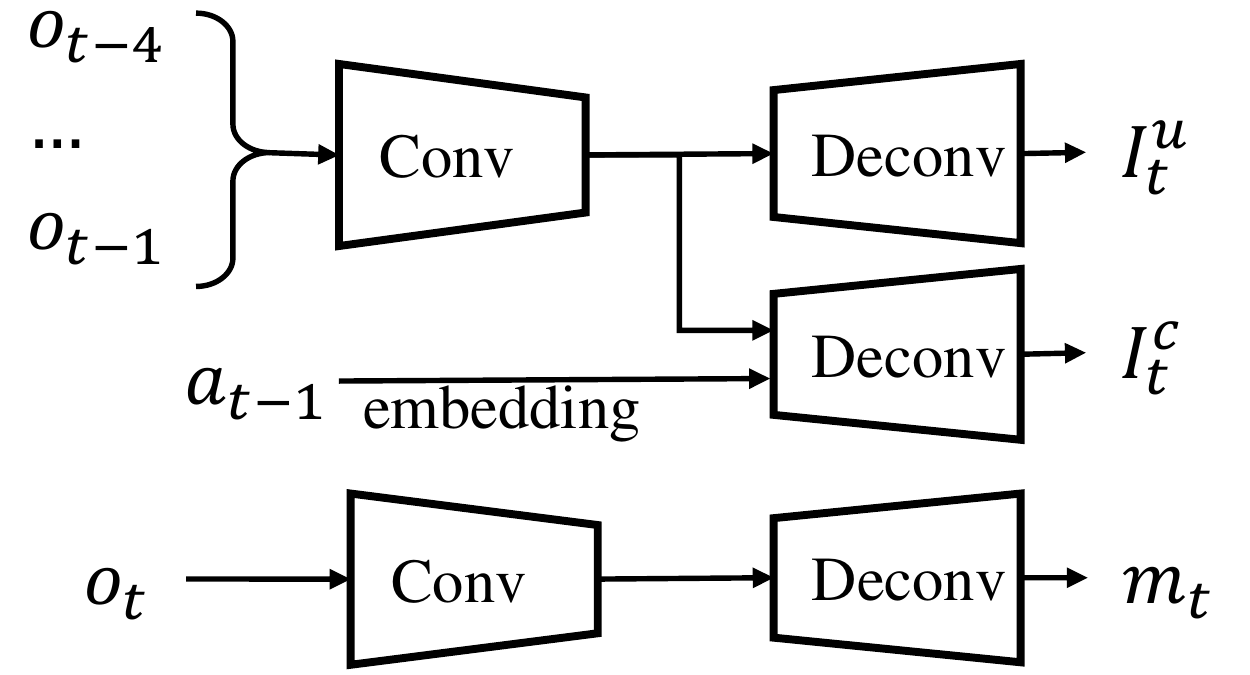}
	\caption{Prediction Network Structure. $o_{t-4},\ldots,o_t$ are frames at time $t-4$ to $t$, $a_{t-1}$ are the action taken after observing $o_{t-1}$. The controllable image part $I^c_t$, the uncontrollable part $I^u_t$ and mask $m_t$ are predicted by three neural net branches.}
	\label{fig_p}
	\vspace{-3mm}
\end{figure}

The detailed structure is shown in Fig.~\ref{fig_p}. The model operates on gray-scale $84 \times 84$ images. The Conv block contains 3 conv layers with 16,32,32 channels, kernel size 6 and stride 2, encoding the previous frames into a $10 \times 10 \times 32$ feature map. The Deconv block is symmetric to the Conv. The embedding for action $a_{t-1}$ has size $10 \times 10 \times 8$ and is merged with the image feature map via a conv layer (kernel size 5 stride 1). The mask prediction $m(o_t)$ has a final sigmoid activation, and everything else has ReLU activation.

\vspace{-10pt}
\subsection{Loss Functions}

We propose the following loss function $L$ (Eq.~\ref{eq_loss_all}) for training,
\begin{align}
	L = &L_{\text{masked}} + L_{\text{recon}} + \lambda_1 L_1 + \lambda_2 L_{\text{act\_pred}} + \lambda_3  L_{\text{flow}}
	\label{eq_loss_all}
\\
	L_{\text{masked}} &= \| m_t \odot o_t - I^c_t \|^2 +  \|  (1-m_t) \odot o_t - I^u_t \|^2
	\label{eq_loss_masked}
\\
	L_{\text{recon}} &= \| o_t - I^c_t - I^u_t \|^2
	\label{eq_loss_sum}
\end{align}
where $\lambda$'s are the coefficients. The notation $\odot$ stands for element-wise product. $\|\cdot\|$ is the $L_2$ norm.

The first two terms are the \emph{main components} of our objective function. Intuitively, since $m$ is expected to take either near 0 or 1 values, $L_{\text{masked}}$ decomposes the target image into non-intersecting parts $m_t\odot o_t$ and $ (1-m_t) \odot o_t$ as training objectives for $I^c_t$ and $I^u_t$ respectively. 
The second term $L_{\text{recon}}$ is the reconstruction error of the whole image. These two terms are consistent, since a perfect model would have $L_{\text{recon}} \approx 0$, $m_t \odot o_t \approx I^c_t$ and $(1-m_t) \odot o_t \approx I^c_t$ all hold.

Since the action information is \emph{only} fed into the $I^c_t$ branch and actions have stochasticity, only the $I^c_t$ branch will learn to account for action-related image changes. And ideally, the $I^u_t$ branch would learn to account for the environment or background changes that are unrelated to the agent's action. The mask $m_t$ balances the two, is implicitly learnt, and forces each pixel to be predicted by either $I^c_t$ or $I^u_t$ but not both.

However, the decomposition into $o_t = I^c_t + I^u_t$ is still quite arbitrary. For instance, a high capacity $I^c_t$ can potentially learn to predict the player's action implicitly and everything else. Moreover, the action related pixel changes do not necessarily correspond to controllable objects, e.g. ghosts chase the PacMan in game MsPacman. Hence we introduce several necessary regularization terms to refine the predictions. 

The $L_1$ term is defined as the $L_1$ regularization on pixel values of $m_t$, $L_1 = \|m_t\|_1$, to encourage a sparser mask.


$L_{\text{act\_pred}}$ is the inverse model regularization term Eq.~\ref{eq_act_pred}, which tries to predict the action between consecutive frames inversely from the mask images of consecutive frames. 
The masks are expected to be refined to focus more on action relevant information. 
$p(\hat{a}_{t-1} = a_{t-1} | m_{t-1},m_{t})$ is parameterized by a simple 2 layer conv net, and $\hat{a}_{t-1}$ is the predicted action, $a_{t-1}$ is the ground truth action.
\begin{equation}
L_{\text{act\_pred}} = -a_{t-1} \log p(\hat{a}_{t-1} = a_{t-1} | m_{t-1},m_{t})
\label{eq_act_pred}
\end{equation}


$L_{\text{flow}}$ is the flow constraint term Eq.~\ref{eq_flow}. This is essentially some smoothness constraint on adjacent masks.
Note in in Atari games, each action has its specific semantic meaning, e.g. left, right, fire, etc. The intuitive idea is, if there is a desired movement in the direction that the action defines, the estimated mask should be continuous in that direction. The $a_x$ and $a_y$ are the desired increments in $x$ and $y$ directions.
\begin{equation}
L_{\text{flow}} = \|m_{t}(x+a_{x},y+a_{y}) - m_{t-1}(x,y)\|^2
\label{eq_flow}
\end{equation}





\section{Application to Deep Q-Learning}

In principle, the prediction model can be readily used alongside any RL algorithms, such as DQN \cite{mnih2015human},  A3C \cite{mnih2016asynchronous}, TRPO \cite{schulman2015trust}, etc. 
Considering deep Q-learning methods already utilize an experience replay buffer, it is convenient to reuse the same buffer to train the prediction model.
During learning, a random minibatch of transitions is sampled every train step to train the prediction model with the aforementioned losses. The Q network is then updated with the usual Bellman loss.

Incorporating the prediction model and using the disentangled image representation as input augmentation to the Double DQN agent \cite{DBLP:conf/aaai/HasseltGS16} yields the \texttt{DDQN+Pred} model. We further show that the prediction model can also generate a useful exploration bonus signal, resulting in the \texttt{DDQN+Pred+Bonus} model.

\begin{figure}[tb]
	\centering
	\includegraphics[width=0.75\linewidth]{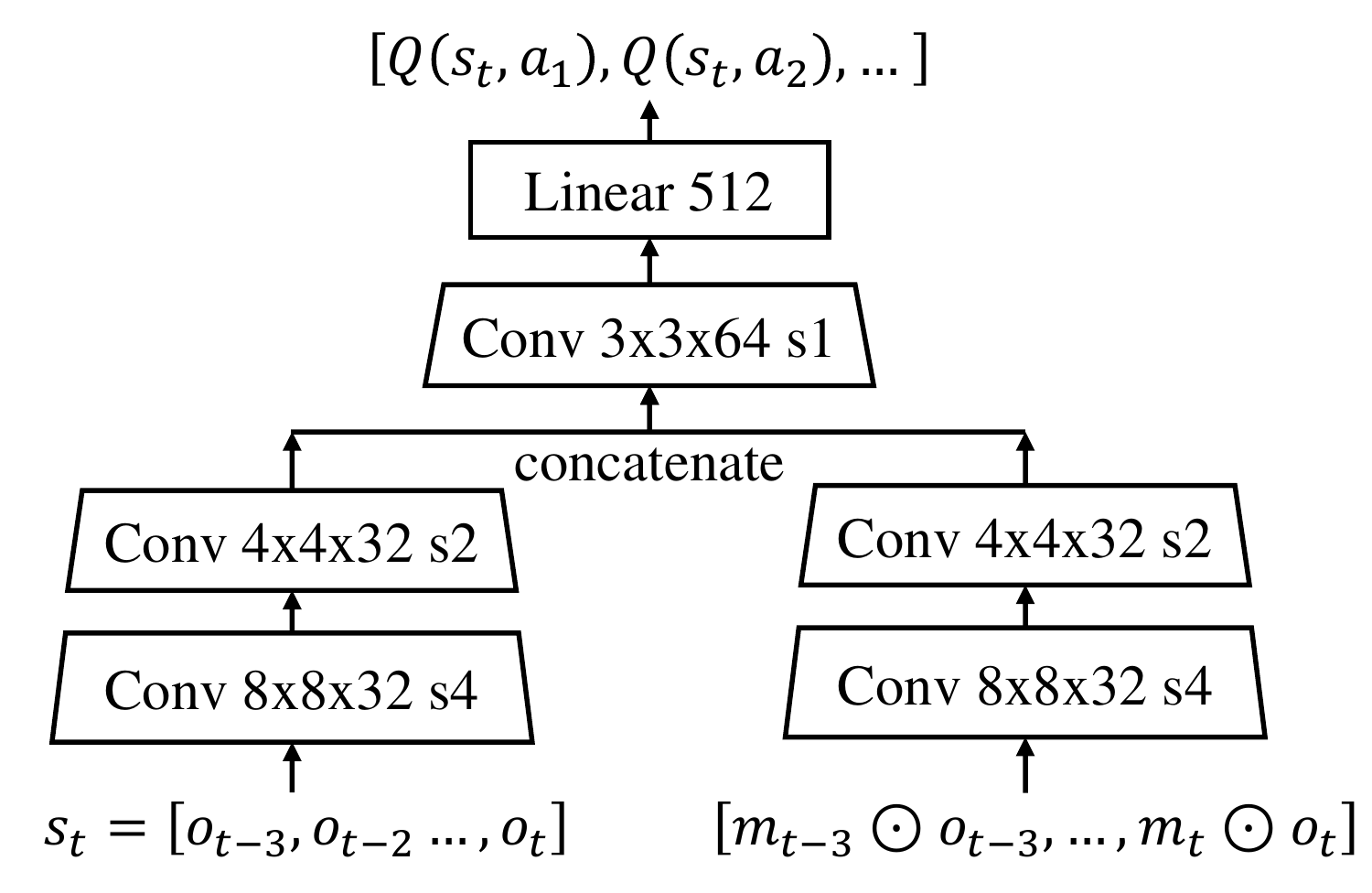}
	\caption{Q Network Structure. The original frames (on the left) and the controllable-object masked images (on the right) are fed into two network streams with \emph{shared} parameters. A late fusion is done before predicting Q values. We mark the conv parameters with $\text{kernel} \times  \text{kernel} \times \text{channels}\;s\,\text{strides} $.}
	\label{fig_q}
	\vspace{-2mm}
\end{figure}




\vspace{-12pt}
\subsection{Augmented Q Network Structure (\texttt{DDQN+Pred})}

In \texttt{DDQN+Pred}, we feed the disentangled representation $m_{t-4+1:t} \odot o_{t-4+1:t}$ as additional input to the Q network. The frame stacking depth, 4, is conventional \cite{mnih2015human}. The modified Q network structure is shown in Fig.~\ref{fig_q}. The representations are merged in a late-fusion manner, as we find it to work better than the early-fusion counterpart. We keep the model size to be comparable to that in the original DDQN by reducing channels by half and sharing parameters of the two streams.




\vspace{-12pt}
\subsection{Prediction Error as Bonus (\texttt{DDQN+Pred+Bonus})}

Following the argument suggested in prior works \cite{stadie2015incentivizing,pathak2017curiosity}, the error of a predictive model can be used as a novelty measure to incentivize exploration during the early stage of learning. At time $t$, the bonus-augmented reward $\Tilde{r}_{t}$ writes as 
\begin{equation}
	\Tilde{r}_{t} = r_t + \frac{\beta}{t} e_t(o_t, a_t) 
	, \;
	e_t(o_t,a_t) = \frac{1}{N_{\text{pix}}} \| m_t \odot o_t - I^c_t \|^2
	\label{eq_bonus}
\end{equation}
where $e_t$ is the prediction error signal defined in Eq.~\ref{eq_bonus}, $N_{\text{pix}}$ is the number of pixels per image, and $\beta$ is a scaling constant.


We choose to include only the action-related part of the prediction error rather than the whole residue $\|o_t - I^c_t - I^u_t\|^2$, mainly because the former contains more information about how well the agent believes it is controlling its avatar. Irrelevant background changes of the frame are suppressed in this way, as those changes might not be beneficial for exploration.

\section{Experimental Results}


\subsection{Qualitative Results of Prediction}



\begin{figure}[bt]
	\centering
	\includegraphics[width=0.9\linewidth]{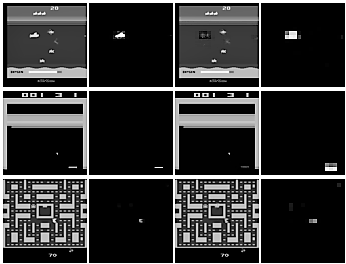}
	\caption{Prediction results for Seaquest, Breakout and MsPacman (rows). The first column: original frames $o_t$, the second column: $I^c_t$, the third column: $I^u_t$, the last column: mask $m_t$. The game frames pre-processed into gray-scale.}
	\label{fig_result_pred}
	\vspace{-2mm}
\end{figure}

Fig.~\ref{fig_result_pred} qualitatively shows the action-related controllable part predictions $I^c$, the action-unrelated uncontrollable part predictions $I^u$, and the predicted masks $m$ that separate out the controllable region, on three games: Seaquest, Breakout and MsPacman. Seaquest is a game in which the player controls a submarine to attack enemies while rescuing divers. The proposed predictive model successfully identifies the submarine from the rest of image as the controllable object. 
Similarly, in Breakout, the predicted mask circles out the paddle on the bottom of the screen that the player can move to catch the ball. 
And in MsPacman, the player controls PacMan to collect rewards on the path and avoid (or attack) enemy ghosts. The predicted mask identifies the PacMan successfully.
In all games, the predictions $I^c$ and $I^u$ make sense in that only the relevant image regions specified by the mask are generated.

\subsection{Agent Performance on Atari Games}

\begin{figure}[t]
	\centering
	\includegraphics[width=\linewidth]{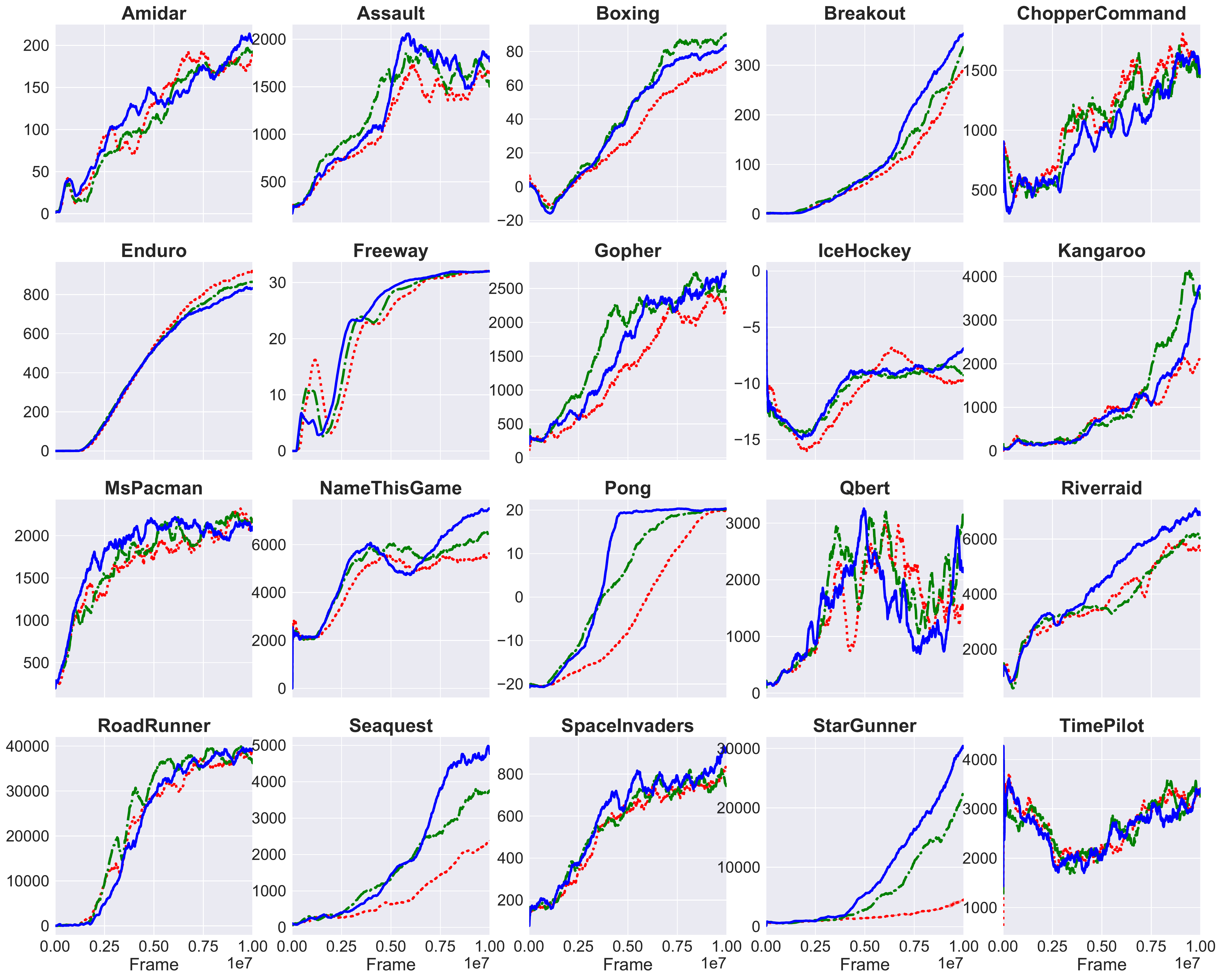}
	\caption{Training curves on Atari games (10 million game frames). Blue lines: \texttt{DDQN+Pred+Bonus} method, green lines: \texttt{DDQN+Pred}, and red lines: DDQN baseline. 
	The plots are averaged over two runs with different random seeds.}
	\label{fig_curve}
	\vspace{-2mm}
\end{figure}

We chose 20 games in total to test the hypothesis that our disentangled representation helps visual RL. Some games have the controllable-avatar property (Breakout, Pong, Seaquest, Freeway, Riverraid, StarGunner, etc.), while some others do not have such property (Enduro, TimePilot, Qbert, IceHockey, ChopperCommand) where no apparent avatar exists or action causes all pixels to change.

We followed the conventional setting of doing gradient update every 4th frame, thus these models are trained for 2.5M gradient steps in total. The hyper-parameters of the baseline DDQN experiments were set to the same as the recommended default values, except for the slightly prolonged decaying of the exploration coefficient $\epsilon$ of the $\epsilon$-greedy policy. The rest of hyper-parameters of our proposed variants were roughly tuned on three games (Breakout, Seaquest, SpaceInvaders), resulting in $\lambda_1 = 0.001, \lambda_2 = 0.1, \lambda_3 = 0.01, \beta = 0.5$. The prediction net was trained by the RMSProp optimizer with learning rate $1\mathrm{e}{-3}$. Batch size was 32. The hyper-parameters were fixed across different games.

The training curves of all methods (DDQN baseline, \texttt{DDQN+Pred} and \texttt{DDQN+Pred+Bonus}) evaluated in Arcade Learning Environments \cite{bellemare13arcade} on the selected games are illustrated in Fig.~\ref{fig_curve}, where the horizontal axis represents time $t$ (number of frames) and the vertical axis represents the episode rewards. The curves are smoothed with a sliding window of 100 episodes. 
In order to obtain fair quantitative comparison, we compute the following normalized score \cite{DBLP:conf/aaai/HasseltGS16},
$$
\text{score}_{normalized} = \frac{\text{score}_{agent} - \text{score}_{random}}{ \text{score}_{human} - \text{score}_{random} } .
$$

As can be seen in Fig.~\ref{fig_curve} and Table~\ref{tbl_atari}, our \texttt{DDQN+Pred} agent outperforms the DDQN baseline on multiple games with the controllable-avatar property including Breakout and Riverraid, and especially Seaquest, Pong and StarGunner, achieving a 252.3\% normalized score with 29.5\% absolute improvement (13\% relative) on average over the baseline. As expected, on the games without the controllable-avatar property, the proposed DDQN variants achieve roughly the same performance as the baseline, indicating that there is no harm utilizing our method. The learning is overall more data efficient with the proposed DDQN variants in many games. For example, in Breakout, Pong and Freeway, the episode scores saturate earlier than the baseline. This could be explained by: Since we already provide the masked image on the controllable object, the agent can treat it as an existing feature instead of learning this feature from scratch, thus can quickly discover and make use of the location of the object in the early training stage.

\begin{table}[t]
	\centering
	\small
	\caption{Performance on Atari games in normalized scores.}
	\begin{tabular}{lll}
		\toprule
		Method & Norm score & Relative gain \\
		\midrule
		Random          & 0\%       & - \\    
		DDQN            & 222.8\%   & 0\% \\   
		DDQN+Pred       & 252.3\%   & +13.2\% \\   
		DDQN+Pred+Bonus & 261.4\%   & +34.5\% \\  
		\bottomrule
	\end{tabular}
	\label{tbl_atari}
	\vspace{-2mm}
\end{table}

On the other hand, the \texttt{DDQN+Pred+Bonus} model which further utilizes the described exploration bonus yields an extra 9.1\% improvement in normalized score over the \texttt{DDQN+Pred} model (in total 38.6\% absolute gain over DDQN). Note that the bonus may not only improve performance in the games with a controllable-avatar, but also in a few other games without an explicit controllable avatar. This could be explained: As long as the predicted mask is not completely empty, the prediction error Eq.~\ref{eq_bonus} is not zero and contains information about the ``familiarity'' of the frame to the prediction net, which may correlate with the visitation count of the agent to that state. 

\section{Related Work}


Action conditioned frame prediction \cite{DBLP:conf/nips/OhGLLS15,leibfried2016deep}, in robotic vision \cite{DBLP:conf/nips/FinnGL16,DBLP:conf/icra/FinnL17}, and in real-world problems \cite{vondrick2016generating}, usually with an encode-transform-decode framework. 
However, the controllable object image regions are not explicitly isolated. Neither do they investigate the possibility of boosting RL with controllable objects.
\cite{DBLP:conf/aaai/BellemareVB12} studies the problem of learning contingent regions, 
whose definition resembles our mask. However, they use the traditional pixel-level transition model, while we adopt an effective deep learning frame prediction method.
The idea of using prediction error as exploration bonus is seen in \cite{pathak2017curiosity,stadie2015incentivizing,DBLP:conf/nips/BellemareSOSSM16}. A recent work \cite{choi2018contingency} studies contingency awareness to promote better exploration for A2C with count-based exploration method, but does not investigate the use of contingency region directly as input or prediction error as bonus, which differs from ours. 
Finally, video prediction model can also be used in model-based RL \cite{DBLP:journals/corr/WeberRRBGRBVHLP17,hafner2019planet,kaiser2019model}. However, predicting images accurately is hard which limits the power of model-based RL, while our method only requires predicting the mask image well.

\section{Conclusions}

We propose a novel learning framework to disentangle the controllable objects from the visual input signal of RL agents. 
We tested our approach on Atari games, and demonstrated improved performance with the controllable object images as additional input and further the prediction error as bonus.

\bibliographystyle{IEEEbib}
\bibliography{refs}

\end{document}